\documentclass[letterpaper]{article} 
\usepackage{aaai24}  
\usepackage{times}  
\usepackage{helvet}  
\usepackage{courier}  
\usepackage[hyphens]{url}  
\usepackage{graphicx} 
\usepackage{natbib}  
\usepackage{caption} 
\usepackage{algorithm}
\usepackage{algorithmic}

\usepackage{newfloat}
\usepackage{listings}

\usepackage{cite}
\usepackage{amsmath}
\usepackage{amsthm}
\usepackage{amssymb}
\usepackage[table]{xcolor}
\usepackage{tikz}
\usepackage{relsize}
\usepackage{appendix}
\usepackage{multirow}
\usepackage{paralist}
\newcommand{\mycircle}[1]{\mathord{\text{\textcircled{\vphantom{X}\ensuremath{#1}}}}}

\DeclareCaptionStyle{ruled}{labelfont=normalfont,labelsep=colon,strut=off} 
\floatstyle{ruled}
\newfloat{listing}{tb}{lst}{}
\floatname{listing}{Listing}
\title{Exploring Self- and Cross-Triplet Correlations for Human-Object Interaction Detection}
\author {
    Weibo Jiang \textsuperscript{\rm 1},
    Weihong Ren \textsuperscript{\rm 1}\thanks{Corresponding author},
    Jiandong Tian \textsuperscript{\rm 2},
    Liangqiong Qu \textsuperscript{\rm 3},
    Zhiyong Wang \textsuperscript{\rm 1},
    Honghai Liu \textsuperscript{\rm 1}
}
\affiliations {
    \textsuperscript{\rm 1} State Key Laboratory of Robotics and System, School of Mechanical Engineering and Automation, Harbin Institute of Technology, Shenzhen\\
    \textsuperscript{\rm 2} State Key Laboratory of Robotics, Shenyang Institute of Automation, Chinese Academy of Science\\
    \textsuperscript{\rm 3} Department of Statistics and Actuarial Science, The University of Hong Kong\\
    
     jiangweibo@stu.hit.edu.cn,
     \{renweihong, wangzhiyong, honghai.liu\}@hit.edu.cn,
     tianjd@sia.cn,
     liangqqu@hku.hk
}

\usepackage{bibentry}

\begin{document}

\maketitle

\begin{abstract}
Human-Object Interaction (HOI) detection plays a vital role
in scene understanding, which aims to predict the HOI triplet
in the form of {\textless human, object, action\textgreater}. Existing methods
mainly extract multi-modal features (e.g., appearance, object semantics, human pose) and then fuse them together to
directly predict HOI triplets. However, most of these methods focus on seeking for self-triplet aggregation, but ignore 
the potential cross-triplet dependencies, resulting in ambiguity of action prediction.
In this
work, we propose to explore Self- and Cross-Triplet Correlations (SCTC) for HOI detection. Specifically, we regard each triplet proposal as a graph where Human, Object represent nodes and Action indicates edge, to aggregate self-triplet correlation. Also, we try to explore cross-triplet dependencies by jointly considering instance-level, semantic-level, and layout-level relations.  
Besides, we leverage the CLIP model to assist our SCTC obtain
interaction-aware feature by knowledge distillation, which provides useful action clues for HOI detection. Extensive experiments on HICO-DET and V-COCO datasets verify the effectiveness of our proposed SCTC.

\end{abstract}

\section{Introduction}

Human-object interaction detection plays a core role in human-centric behavior analysis tasks and provides deeply understanding of human intentions and behaviors~\cite{chao2015hico,gupta2015vcoco}. HOI detection approaches need to extract the semantic relationships between humans and objects and finally predict a set of {\textless human, object, action\textgreater} triplets within an image. The challenges in the HOI detection task mainly lie in the modeling the relationships between human and object and leveraging these relationships to perform human-object paring as well as predict their interactions. 

The HOI detection approaches can be roughly divided into two categories: one-stage detectors and two-stage ones.
One-stage methods implicitly build the relationship between human and object, and try to simultaneously perform object detection and interaction prediction. 
E.g., previous CNN-based work~\cite{wang2020IPNet} localizes and classifies the interaction by detecting additional interaction point, while other method~\cite{kim2020uniondet} directly captures the interaction region using a union-level box. 
Though the CNN-based methods can effectively model human-object relationships, they usually need post-grouping strategies to form a complete triplet.   
Recently, to capture long range contexts, the transformer-based works~\cite{tamura2021qpic,zou2021end,dong2021visual,kim2022mstr,liao2022gen-vlkt,zhou2022disentangled,kim2023relational} have greatly advanced the HOI detection using self-attention and cross-attention mechanisms.
They directly predict a HOI triplet without extracting instance-level priors or exploring their dependencies, but this end-to-end way may suffer insufficient exchange between contextual clues, potentially leading to a sub-optimal solution.

\begin{figure}
    \centering
    \includegraphics[width=1.0\linewidth]{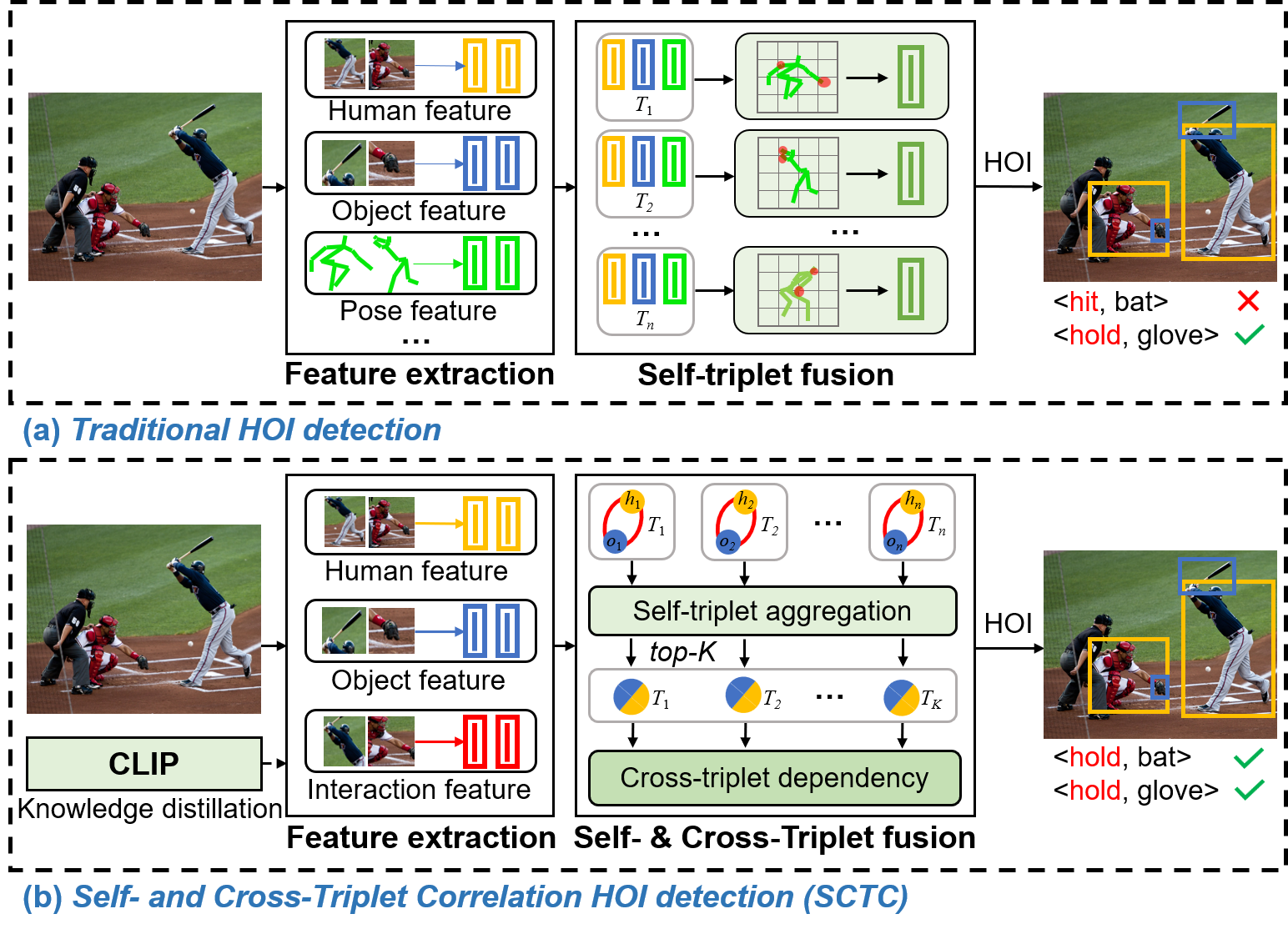}
    \caption{Architecture comparison of different HOI detection pipelines. (a) Traditional HOI detection methods focus on exploring multiple features and fuse them by self-triplet fusion (e.g. body-part attention); 
    (b) our SCTC jointly explores Self- and Cross-triplet Correlations, aiming to eliminate action ambiguity and promote scene understanding. 
    }
    \label{fig:architecture comparation}
    \vspace{-0.25in}
\end{figure} 

The two-stage methods firstly perform object detection, and then explicitly build the relationships between human and object to perform HOI prediction. As shown in Fig.~\ref{fig:architecture comparation} (a), some works ~\cite{xu2019learning,kim2020cooccur} extract multi-modal features (e.g., appearance, layout, or human pose) based on object detection, and then fuse them through the Multi-Layer Perceptron (MLP) for final action prediction. However, such simple
fusion fails to delve deep correlations between different instance-level features.
The recent works \cite{li2022transferable,park2023viplo} jointly use instance level and body part features to mine interactivenesses in HOI. However, they only focus on seeking for the self-triplet attentions, but ignore the potential cross-triplet dependencies, which may lead to ambiguous action prediction. 
In contrast, as shown in Fig.~\ref{fig:architecture comparation}(b), we propose to explore Self- and Cross-Triplet Correlations (SCTC) for HOI detection. Using object detections, we first extract instance-level and interaction-level features simultaneously.
Then, we regard each candidate triplet as a graph (``human" and ``object" are nodes, ``interaction" is edge) to aggregate self-triplet features. After that, we model a triplet-level graph where each node represents a triplet proposal, to explore cross-triplet dependencies. Here, the instance-level, semantic-level, and layout-level relations are used to formulate adjacency matrix, thereby eliminating ambiguitious action prediction.

Interaction features play a vital role in self-triplet aggregation, and they are  utilized to associate different humans and objects, which helps to predict the action of a human-object pair. To enhance this representation,
we leverage the CLIP model~\cite{radford2021clip} to transfer its vision-language knowledge to the interaction feature. This is different from other methods~\cite{liao2022gen-vlkt, yuan2022rlip,ning2023hoiclip,wan2023weakly}, which adopt CLIP to distill the HOI classification in the final stage for few-shot detection.
Experimental results on public datasets show the superiority of our proposed model. Also, the ablation studies prove the effectiveness of the Self- and Cross-Correlation mechanism. 

To summarize, our contributions are three-fold:
\begin{itemize}
    \item In order to obtain self-triplet aggregation, we formulate a graph for each triplet where human and object are represented as nodes, and the interaction prior is innovatively employed as the graph edge that models relationships between different instances.
    \item To build the contextual dependencies across different candidate triplets, we also create a triplet-level graph, and utilize the instance-level, semantic-level, and layout-level relations to formulate adjacency matrix, thereby eliminating ambiguous action prediction.
    \item We adopt CLIP model to transfer vision-language knowledge to the interaction feature, which serves as the key element in exploring the semantic relationships from different instances.
\end{itemize}

\section{Related Works}
\subsection{One-Stage HOI Detection Methods}
The one-stage HOI detection methods predict the HOI triplet directly, and build the dependencies between human and object using geometric or structural constrains.
Previous CNN-based works~\cite{kim2020uniondet,liao2020ppdm,wang2020IPNet,zhong2021glance} predict the triplet using two parallel branches: in the first branch, the image feature map is used to find the interaction point or the interaction area and accordingly the interaction.
In the second branch, object detection is performed to find the human-object pair. Finally, the two branches are used to regress the offsets and match the interaction class with the detected instances.
These methods capture the contextual information by pairing the human and object from an early stage in feature extraction. However, when the interacting human and object are far away from each other, the limited global modeling capability of CNNs may lead to ambiguity in the semantic features.

Recent transformer-based works~\cite{zou2021end,tamura2021qpic,kim2021hotr,zhou2022disentangled,liao2022gen-vlkt,lim2023ernet} leverage the encoder-decoder architecture to jointly represent human, object, and interaction features, and build their relationships implicitly.
To learn the representations that focus on different feature regions, Kim \emph{et al.}~\shortcite{kim2023relational} propose three decoder branches to represent human, object, and interaction, respectively.
The relationships are built through large amount of self-attention and cross-attention modules, leading to the increase of computation cost.
These methods made great exploration of applying transformer to HOI detection, however, they only build associations of HOI in instance level, and also suffer from the low training speed and large memory usage~\cite{park2023viplo}.

\begin{figure*}[h]
    \centering
    \includegraphics[width=0.95\linewidth]{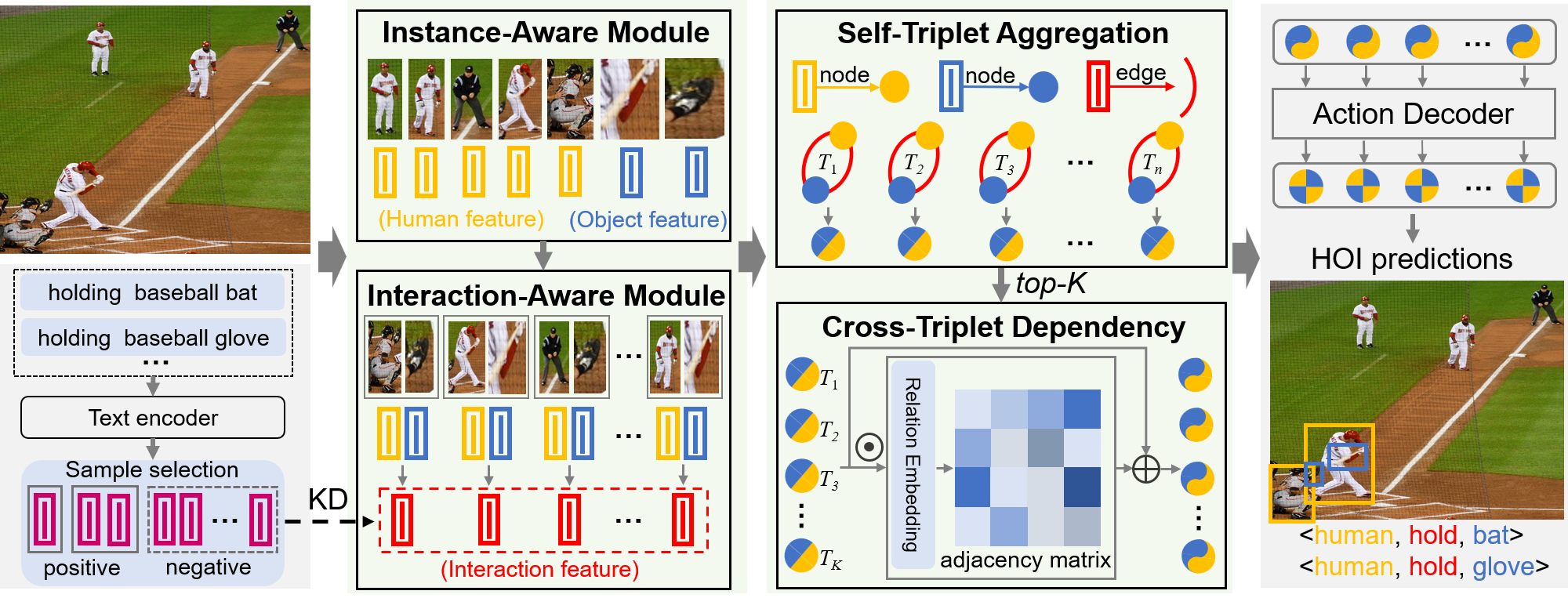}
    \caption{The overall pipeline of SCTC.
    For an input image, the instance-aware module is first used to extract instance-level features (appearance and semantics).
    Then, the interaction-aware module match human-object pairs and fuse them together to generate  interaction feature. 
    Next, Self-Triplet Aggregation (STA) is employed to explore self-triplet attentions, while the Cross-Triplet Dependency (CTD) is used to build connections across different HOI triplets.
    Finally, the action decoder is utilized to predict HOI triplets. 
    Also, Knowledge Distillation (KD) is used to transfer text embeddings from CLIP to interaction feature.  
    }
    \label{fig:SCTC}
    \vspace{-0.10in}
\end{figure*} 

\subsection{Two-Stage HOI Detection Methods}
The two-stage HOI detection approaches initially detect all possible interacting pairs,
and then extract the corresponding features (e.g., appearance, spatial, pose, semantics).  
Finally, these features are fused to predict the interactiveness and the action for each pair.
Chao \emph{et al.}~\shortcite{chao2018learning} propose the multi-stream framework to 
obtain contextual information for interaction classification, but it only considers the spatial relations. 
The subsequent works encode a variety of additional priors for robust representation, including pose features~\cite{kim2020cooccur,li2022transferable,park2023viplo} and semantic features~\cite{bansal2020detecting,liu2022highlighting,zhang2022exploring}.
E.g., Zhang \emph{et al.}~\shortcite{zhang2022exploring} and Kim \emph{et al.}~\shortcite{kim2020cooccur} first concatenate multi-modal features and then use the MLP to encode them for interaction classification, but such 
simple fusion strategy fails to mine the relationships between these features.
Li~\emph{et al.}~\shortcite{li2022transferable} explore the interactiveness using human pose, and build the dependencies between human and object based on the human-part attentions. However, it only considers the self-triplet relationships, which may lead to action ambiguity.

Recent studies~\cite{gao2020drg,zhou2021cascaded,zhang2021spatially,park2023viplo} leverage graph network to explore relationships between human and object.
E.g., Park~\emph{et al.}~\shortcite{park2023viplo} employ human joints to focus on specific human regions, and then use a graph network to aggregate local features within each human-object pair.
The above graph-assisted methods have made contributions to explore the potential relationships between human and object.
However, they only model the instance-level relations within a HOI triplet, and don't consider the dependencies across different triplets.
Thus, in this work, we propose to explore self- and cross-triplet correlations in HOI detection, and encode the interaction feature as the fundamental element for the reasoning process, which is a new attempt in HOI relationship exploration.

\subsection{Exploiting Vision-Language Models}
Recently, the CLIP model~\cite{radford2021clip} has demonstrated strong generalization to various downstream tasks~\cite{gu2022open,li2022language,esmaeilpour2022zero}, and several studies have attempted to leverage the knowledge from CLIP to HOI detection.
E.g., Dong \emph{et al.}~\shortcite{dong2022category}, Yuan \emph{et al.}~\shortcite{yuan2022rlip}, Wan \emph{et al.}~\shortcite{wan2023weakly} and Zhang \emph{et al.}~\shortcite{zhang2023sqa} integrate text embeddings generated from the CLIP to enhance the representation of semantic features, and have achieved promising results.
In contrast, Liao \emph{et al.}~\shortcite{liao2022gen-vlkt} leverage the CLIP knowledge for interaction classification and visual feature distillation, but such a kind of distillation may introduce ambiguity when multiple HOI triplets exist in the scenario.
Ning \emph{et al.}~\shortcite{ning2023hoiclip} also try to transfer visual-linguistic knowledge to HOI detection, utilizing the prior knowledge in CLIP to generate a classifier by encoding HOI descriptions.
Though this method can effectively integrate CLIP knowledge, it still requires the participation of CLIP in the 
inference stage, which results in a high computation cost.
Different from the methods above, we leverage the CLIP model to distill the interaction feature existing in each {\textless human, object, action\textgreater} triplet, which is a fine-grained knowledge distillation method.

\section{Method}
We propose the Self- and Cross-Triplet Correlation HOI model to explore relationships among humans, objects, and interactions for better understanding the human-object pairing and the following action prediction.
As shown in Fig.~\ref{fig:SCTC}, we first leverage Instance-Aware Module to detect all the objects and obtain the instance-level features.
Then, we employ Interaction-Aware Module to perform human-object pairing and project the fused features into interaction feature.
After that, the Self-Triplet Aggregation (STA) is adopted to explore self-triplet attentions, while the following Cross-Triplet Dependency (CTD) is used to build connections across different HOI triplets. 
Finally, the actions for human-object pairs are predicted through the action decoder.

\subsection{Instance-Aware Module}
\label{Insmodule}
Given an input image, we first leverage the DETR~\cite{carion2020detr}, which is composed of ResNet~\cite{he2016resnet} and the following Transformer~\cite{vaswani2017attention}, to detect all the humans and objects. Then, we extract the features of them from two aspects: appearance and semantics. The appearance feature is directly derived from the output of DETR decoder, which is highly related with the detected object. The semantic feature is obtained by embedding the object category into a feature vector. Finally, both the human feature ${\bf F}_{h}$ and the object feature ${\bf F}_{o}$ can be represented as the concatenation of appearance and semantic feature.

\subsection{Interaction-Aware Module}
\label{Intmodule}
The interaction-aware module aims to learn the interaction feature from the instance-level features (${\bf F}_{h}$ and ${\bf F}_{o}$). Specifically, we pair all the detected humans and objects to construct candidate human-object pairs.
The hard mining strategy~\cite{shrivastava2016training} is employed to sample negative pairs to facilitate the learning of interactiveness prediction following Zhang~\emph{et al.}~\shortcite{zhang2022exploring}.
Then, we extract the spatial feature of each candidate pair, which can be denoted as $\left[ \Delta x,\Delta y,\Delta s,\arctan (\frac{\Delta y}{\Delta x}),{{A}_{h}},{{A}_{o}},{{A}_{I}},{{A}_{U}} \right]$, where $\Delta x$, $\Delta y$, $\Delta s$  denote the distance between human and object in the $x$-axis, $y$-axis, and the coordinate system, respectively; ${A}_{h}$ and ${A}_{o}$ denote the area of human and object, respectively; ${A}_{I}$ and ${A}_{U}$ denote the intersection area and union area of the human-object pair, respectively.
By concatenating the human feature ${\bf F}_{h}$, object feature ${\bf F}_{o}$, and the spatial feature together, we obtain the pair-level feature ${\bf F}_{ho}$, which contains the appearance, semantic, and spatial features of a human-object pair.
Finally, the interaction feature ${\bf F}$ of each pair is obtained by projecting its corresponding ${\bf F}_{ho}$ with the MLP.

\subsection{Visual-Linguistic Knowledge Transfer}
To enhance the interaction feature, we transfer the visual-linguistic knowledge from CLIP~\cite{radford2021clip} to our interaction-aware module. Specifically, we first adopt the CLIP model to generate text embeddings for the ground truth HOIs. Next, we utilize the text embeddings to distill our interaction feature ${\bf F}$.

To generate the text embeddings, we first convert the HOI triplet labels into textual descriptions.
Given an object and a verb, we use the template of ``A photo of a person [verb-ing] a/an [object]'' to generate a sentence.
Besides, the verb of ``non-interaction'' is represented as ``A photo of a person have non interaction with the [object]''.
Then we generate the text embedding ${\bf E}$ for each candidate human-object pair through the pre-trained CLIP text encoder offline. 

After that, we leverage the text embeddings to distill our interaction feature.
For positive human-object pairs that involve a single interaction,
we directly take their corresponding text embeddings as supervision, which promotes the interaction feature learning from the text encoder of CLIP.
While for other positive pairs with multiple interactions, we calculate the average representation of the text embeddings for each pair to minimize semantic ambiguity.
Besides, for the negative pairs, we take the average representation of all the text embeddings, which helps to push interaction feature away from non-interactive cluster.
Formally, we use the $\mathcal{L}_{1}$ loss to supervise the distance between the interaction feature ${\bf F}$ and the text embedding ${\bf E}$: 
\begin{equation}
\label{eq:kd}
{{{\mathcal{L}}_{kd}}=\frac{1}{{{N}_{p}}}\sum\limits_{i}{\left| {{{{\bf {E}}_{i}}}}-{{\bf{F}}_{i}} \right|}},
\end{equation}
where ${i}$ denotes the $i$-th candidate human-object pair, and ${N}_{p}$ is the number of candidate human-object pairs.

\subsection{Self-Triplet Aggregation
\label{Intramodule}}
Previous works explore self-triplet relations solely on the basis of instance-level features, but ignore the interaction feature, which is the fundamental element of a HOI triplet. 
In our Self-Triplet Aggregation (STA), we propose to aggregate features from both instance level and interaction level.
Specifically, we design a graph neural network~\cite{kipf2016graph} for each candidate triplet, 
where the human and object are viewed as nodes and the interaction feature is treated as edge.
The human node ${{\nu }_{h}}$ and object node ${{\nu }_{o}}$ are directly derived from the human feature ${\bf F}_{h}$ and object feature ${\bf F}_{o}$, respectively. 
Each node (${{\nu }_{h}}$ or ${{\nu }_{o}}$) contains the appearance feature and semantic feature for a specific instance, which only focuses on the instance itself.
To aggregate features from ${{\nu }_{h}}$ and ${{\nu }_{o}}$, we initialize the edge $\varepsilon$ as the  interaction feature ${\bf F}$.
The edge contains the interaction semantics, spatial feature, and instance features, forging a robust connection between human and object in a candidate HOI pair.

We aim to build correlation between human and object via the interaction-aware edge, empowering STA to capture both individual attributes and the potential relationship between them.
For the positive HOI triplets, the interaction feature distilled by CLIP tends to encompass specific action semantics, steering the human node ${{\nu }_{h}}$ and object node ${{\nu }_{o}}$ to acquire action clues through the edge $\varepsilon$.
Meanwhile, for the negative triplets, the interaction feature learns the semantics of non-interactiveness, and the implicit information that the human and object is not a interactive pair is transmitted to the corresponding nodes. 

The human node and the object node are updated with the message propagation from the each other and the edge:
\begin{equation}
\label{eq:h2o}
{{\nu }_{h\to o}}={{f}_{i}}(\varepsilon )\odot {{f}_{h}}({{\nu }_{h}}),
\end{equation}
\begin{equation}
\label{eq:o2h}
{{\nu }_{o\to h}}={{f}_{i}}(\varepsilon )\odot {{f}_{o}}({{\nu }_{o}}),
\end{equation}
where ${{f}_{i}}$, ${{f}_{h}}$, ${{f}_{o}}$ are projection functions consist of a linear function and a ReLU activation function, and $\odot$ is the Hadamard product.

The updated nodes can be obtained by:
\begin{equation}
\label{nno}
{{\widehat{\nu }}_{o}}={{\nu }_{h\to o}}+{{\nu }_{o}}, 
\end{equation}
\begin{equation}
\label{nno}
{{\widehat{\nu }}_{h}}={{\nu }_{o\to h}}+{{\nu }_{h}}.
\end{equation}
Here, we use skip connections to enable the graph network to learn the additional correlations between human node and object node. After message propagating, each node has a comprehensive understanding of its potential interactions.

Afterwards, we concatenate the updated human and object nodes together (denoted as ${\nu}_{hoi}$) to calculate the interactiveness scores followed by two-layer linear projections. We select the top-K human-object proposals according to the interactiveness scores, and the corresponding features are also abstracted.
We optimize the human-object matching process by focal loss (FL)~\cite{lin2017focal}:
\begin{equation}
\label{hopair}
{{{\mathcal{L}}_{pair}}=\frac{1}{{{\sum\nolimits_{k=1}^{K}{p}}_{k}}}\sum\limits_{k=1}^{K}{{\bf FL}({{\widehat{p}}_{k}},~}{{p}_{k}})},
\end{equation}
where ${{p}_{k}}\in \left\{ 0,1 \right\}$ indicates whether an interaction exists in ground truth and ${{\widehat{p}}_{k}}$ is the predicted interactiveness score.

\subsection{Cross-Triplet Dependency}
\label{Intermodule}
The current two-stage works focus on exploring relationships in the self-triplet level, but ignore the implicit dependencies in the cross-triplet level. 
Actually, the relations between humans and objects are often context-dependent and can affect the reasoning of individual interactions.
Thus, we create the Cross-Triplet Dependency (CTD), a triplet-level graph to exploit the correlations across different triplets, for eliminating action ambiguity and complete scene understanding.
Different from STA, each node in CTD denotes a HOI triplet proposal and is initialized by the ${\nu}_{hoi}$ obtained in STA. Here, we mainly introduce the formulation of the edges, i.e., the Adjacency Matrix.

\begin{figure}
    \centering
    \includegraphics[width=0.90\linewidth]{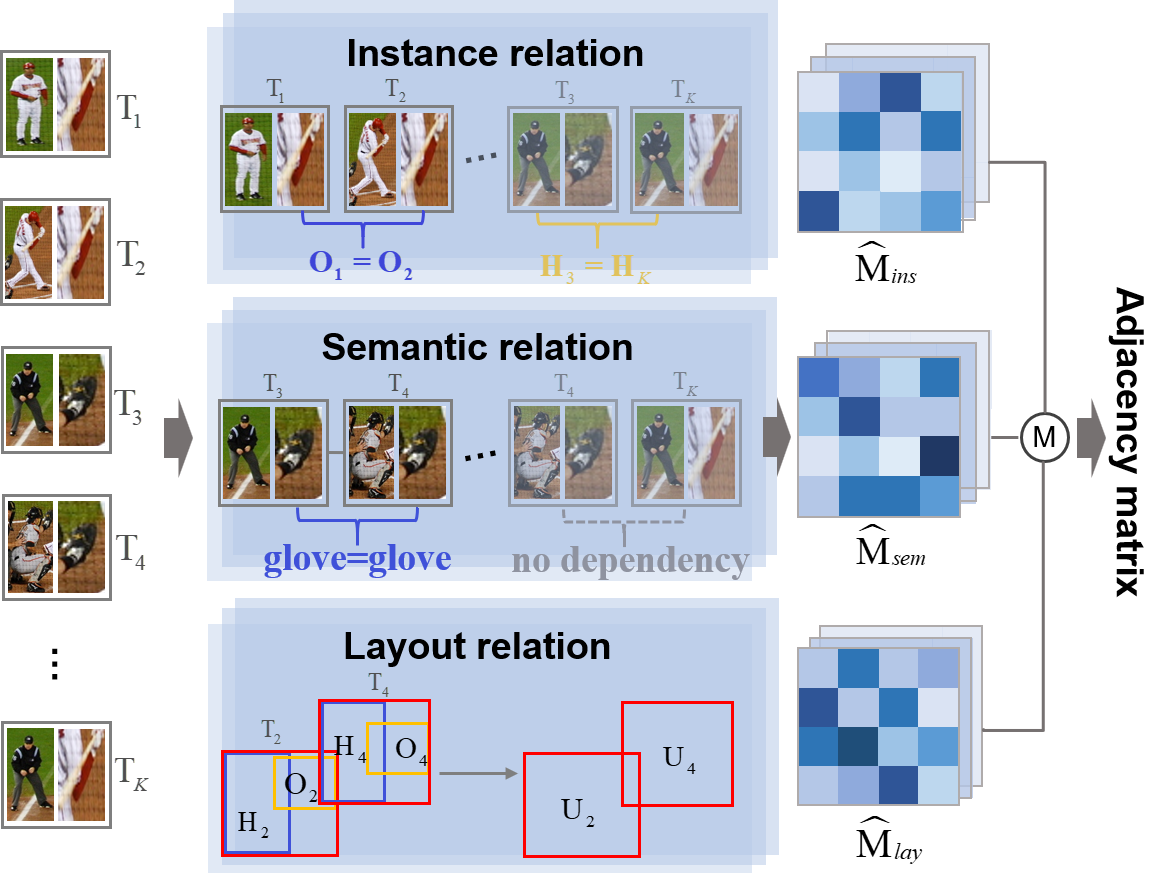}
    \caption{Generation of the edges in CTD, where the $ \mycircle{\text {\scriptsize M}}$ indicates the MLP.}
    \label{fig:cross-triplet}
    \vspace{-0.25in}
\end{figure} 

As shown in Fig.~\ref{fig:cross-triplet}, we embed the instance relation, semantic relation, and layout relation into an adjacency matrix to link all the nodes.
The instance relation means whether different HOI triplets share the same human or object. We create a matrix ${{\bf M}_{ins}}\in {{\mathbb{R}}^{K\times K\times 2}}$ to represent the instance relation, where $K$ means the selected $K$ triplets.
If the $j$-th triplet ${\nu}_{hoi}^{j}$ and the $k$-th triplet ${\nu}_{hoi}^{k}$ share the same human or object, the corresponding value ${{\bf M}_{ins}^{(j,k,0)}}$ or ${{\bf M}_{ins}^{(j,k,1)}}$ is set to 1 (channel ``0" and ``1" represent human and object, respectively).
Otherwise, the value is set to 0.
If there exists the instance-level connection between two different triplets, there maybe exists the interaction relationship, which is helpful for the HOI detection with multiple actions. 
For example, if we find a triplet of {\textless human, hold, fork\textgreater}, it is high likely that the mentioned human is also associated with another triplet of {\textless human, hold, knife\textgreater}.
The instance relation aims to exploit such kind of implied relations, and its embedding ${{\widehat{\bf M}_{ins}}}\in {{\mathbb{R}}^{K\times K\times {{d}_{ins}}}}$ (${d}_{ins}=64$) is obtained by the linear embedding on ${{\bf M}_{ins}}$.

The instance relation only considers whether the instances in different triplets are identical,
neglecting the dependencies among different triplets with the same object category, and may cause a lack of semantic information.
Thus, the semantic relation is introduced  
to measure whether different HOI triplets share the same object semantics. 
We design a matrix ${{\bf M}_{sem}}\in {{\mathbb{R}}^{K\times K\times {{C}_{o}}}}$ to represent the semantic relation, where ${C}_{o}$ denotes the number of object categories. If ${\nu}_{hoi}^{j}$ and ${\nu}_{hoi}^{k}$ share the same object category $c$, the corresponding value ${{\bf M}_{sem}^{(j,k,c)}}$ is set to 1.
Otherwise, the value is set to 0.
For example, if we find a triplet of {\textless human, ride, motorbike\textgreater}, it is likely that another triplet with the same interaction also exists in the scenario but hard to be detected.
The semantic relation aims to inject the semantic information to the adjacency matrix and find the hard positive samples, and its embedding ${{\widehat{\bf M}_{sem}}}\in {{\mathbb{R}}^{K\times K\times {{d}_{sem}}}}$ (${d}_{sem}=256$) is obtained by the linear embedding on ${{\bf M}_{sem}}$.

The layout relation aims to build spatial relationships between two different triplets. We first calculate the union box for each triplet, and then 
extract the spatial feature between the two union boxes, which can be represented as $\left[ \Delta x,\Delta y,\Delta s,\arctan (\frac{\Delta y}{\Delta x}),{{A}_{h}},{{A}_{o}},{{A}_{I}},{{A}_{U}} \right]$ following the interaction-aware module.
Thus, the layout relation ${{\bf M}_{lay}}\in {{\mathbb{R}}^{K\times K\times 8}}$ is formulated by calculating the spatial feature between every two triplets. The corresponding embedding ${\widehat{\bf M}_{lay}}\in {{\mathbb{R}}^{K\times K\times {d_{lay}}}}$ (${d}_{lay}=64$) is also obtained by the linear embedding on ${\bf M}_{lay}$.

After all the relations are built and embedded, the final adjacency matrix is derived by:
\begin{equation}
\label{adj}
{{\bf M}_{adj}}=\operatorname{\bf MLP}({\bf Concat}({{\widehat{\bf M}}_{ins}},{{\widehat{\bf M}}_{sem}},{{\widehat{\bf M}}_{lay}})),
\end{equation}
where ${\bf Concat}$ is the concatenation, and ${{\bf M}_{adj}}\in {{\mathbb{R}}^{K\times K}}$.
Then, we update the HOI triplet nodes by:
\begin{equation}
\label{inter_pass}
{{\widehat{\nu }}_{hoi}}=({{{\bf M}}_{adj}}\cdot {{\nu }_{hoi}})\oplus {{\nu }_{hoi}}
\end{equation}
where $\cdot$ is the matrix multiplication, and $\oplus$ is the element-wise addition.
Similar to the STA, we still adopt the skip connections to enable the CTD to learn additional cross-triplet dependencies.
Through (\ref{inter_pass}), the correlations 
across different triplets are established for final action prediction.

\subsection{Human-Object Interaction Prediction}
As shown in Fig.~\ref{fig:SCTC}, we predict the final actions for each human-object pair using the action decoder.
The decoder query is initialized by projecting the ${\widehat{\nu }}_{hoi}$ with the MLP, and the key and value come from the features extracted by the ResNet50 in DETR.
Each query is well refined with STA, which integrates interaction semantics between human and object.
Also, CTD builds the dependencies across different queries for further exploring action clues.
The final actions $\widehat{\bf y}$ are obtained by linear projection on the output of action decoder, and the classification is optimized by focal loss:
\begin{equation}
\label{final action}
{{{\mathcal{L}}_{a}}=\frac{1}{\sum\nolimits_{k=1}^{K}{\sum\nolimits_{c=1}^{C}{{{\bf y}_{k,c}}}}}\sum\limits_{k=1}^{K}{\sum\limits_{c=1}^{C}{{\bf FL}({{\widehat{\bf y}}_{k,c}},~}}{{\bf y}_{k,c}})},
\end{equation}
where ${{\bf y}_{k,c}} \in \left\{ 0,1 \right\}$ indicates whether the $k^{th}$ proposal contains the $c^{th}$ action class, and ${{\widehat{\bf y}}_{k,c}}$ is the predicted probability of $c^{th}$ action occurs in the $k^{th}$ proposal.

Finally, the HOI predictions are achieved by combining the human-object pairs and the predicted actions.
The overall objective of our SCTC contains the knowledge distillation in~(\ref{eq:kd}), the human-object proposal prediction in~(\ref{hopair}), and the action classification in~(\ref{final action}):
\begin{equation}
\label{all loss}
{{\mathcal{L}}_{SCTC}}=
\alpha \cdot {\mathcal{L}_{kd}}+
\beta \cdot {{\mathcal{L}}_{pair}}+
\gamma \cdot {{\mathcal{L}}_{a}},
\end{equation}
where $\alpha, \beta, \gamma$ are hyper-parameters.

\section{Experiment}
We evaluate our SCTC on two commonly used HOI detection datasets: HICO-DET~\cite{chao2015hico} and V-COCO~\cite{gupta2015vcoco} .

\subsection{Experimental Setting}
\subsubsection{Datasets.}
V-COCO is a subset of MS-COCO~\cite{lin2014microsoft}, consisting of 5,400 images in the trainval set and 4,946 images in the test set.
It has 259 HOI categories over 29 actions and 80 objects. 
HICO-DET consists of 38,118 images in training set and 9,658 in test set, and has 600 HOI categories over 117 actions and 80 objects. The 600 HOI categories are split into 138 Rare and 462 Non-Rare based on the number of instances.

\subsubsection{Evaluation metric.}
Following the standard metric, we use the commonly used mean Average Precision (mAP) to evaluate the model performance for the two benchmarks.
A HOI triplet is considered as true positive if it localizes the human and object accurately (i.e., the Interaction-over-Union (IOU) between the predicted bounding boxes and ground truth is greater than 0.5) and also predicts the action correctly.

\subsubsection{Implementation details.}
We take the DETR for object detection.
All the learnable parameters in DETR are frozen during training. We leverage the pre-trained ViT-16/B CLIP on its official data to generate text embeddings for knowledge distillation, and also keep its parameters fixed during training. 
We set the number of action decoder layers to 6 and the number of queries to 32, which means that we select the top-32 human-object proposals in the Self-Triplet Aggregation (STA). 
The parameters $\alpha $, $\beta $, $\delta $ are all set to 1.0. 
The entire model is trained on two Nvidia 3090 GPUs with a batch size 8. The AdamW~\cite{loshchilov2017AdamW}
optimizer is used for training with 30 epochs, where the
starting learning rate is $5\times {{10}^{-5}}$, and then decays with the Cosine annealing training strategy. 

\subsection{Comparisons with the State-of-the-Arts}
\begin{table}
\begin{center}
\setlength{\abovecaptionskip}{0.05 cm} 
\setlength{\belowcaptionskip}{-0.05 cm} 
\caption{Performance comparison on HICO-DET dataset.
The best result is marked with bold and the second best result is underlined.
For fair comparison, we report results using an object detector finetuned on the training dataset of HICO-DET for the two-stage methods.
}
\label{table:HICO-DET}
\resizebox{\linewidth}{!}
{
\begin{tabular}{llccc}
\hline
Method & Backbone  & Full & Rare & Non-Rare \\
\hline\hline
\emph{One-stage methods} &&&\\
UnionDet~\cite{kim2020uniondet} & R50  &14.25 &10.23 &15.46 \\
IPNet~\cite{wang2020IPNet} & HG-104  & 19.56 & 12.79 & 21.58 \\
HOTR~\cite{kim2021hotr} & R50 & 23.46 & 16.21 &25.60 \\
AS-Net~\cite{chen2021reformulating} & R50 & 24.40 & 22.39 &25.01 \\
HOITrans~\cite{zou2021end} & R101 & 26.61 & 19.15 & 28.84 \\
QPIC~(Tamura et al. 2021) & R101 & 29.90 & 23.92 &31.69\\
PhraseHOI~\cite{li2022improving} & R101 & 30.03 & 23.48 & 31.99\\
Zhou~\cite{zhou2022disentangled} & R50 & 31.75 & 27.45 & 33.03 \\
CATN~\cite{dong2022category} & R50 & 31.86 &25.15 & 33.84 \\
CDN~\cite{zhang2021mining} & R101 & 32.07 & 27.19 & 33.53 \\
GEN-VLKT~\cite{liao2022gen-vlkt} & R50  & 33.75 & 29.25 & 35.10 \\
HOICLIP~\cite{ning2023hoiclip} & R50 & 34.69 & 31.12 & 35.74\\
ParMap~\cite{wu2022mining} &R50 &35.15 &33.71 &35.58\\
ERNet~\cite{lim2023ernet} & EfficientNetV2-L & 35.92 & 30.12 & {38.29} \\
\hline
\emph{Two-stage methods} &&&\\
InteractNet~\cite{gkioxari2018detecting} & R50-FPN & 9.94 & 7.16 & 10.77  \\
GPNN~\cite{zhou2021cascaded} & R50 & 19.42 & 13.98 & 20.91\\
ACP~\cite{kim2020cooccur} &R152 & 20.59 &15.92 &21.98\\
OCN~\cite{yuan2022detecting} & R101 & 31.43 & 25.80 & 33.11\\
STIP~\cite{zhang2022exploring} & R50 & 32.22 & 28.15 &33.43 \\
UPT~(Zhang et al. 2022a) & R101 & 32.62 & 28.62 & 33.81 \\
OpenCat~(Zheng et al. 2023) & R101 & 32.68 & 28.42 & 33.75\\
CQL~\cite{xie2023category} & R101 & 36.03 & 33.16 & 36.89\\
ViPLO~(Park et al. 2023) & ViT-B/16 & {37.22} & {35.45} & 37.75\\
RmLR~\cite{cao2023re} & R101 & {37.41} & {28.81} & {\bf 39.97}\\
\rowcolor{lightgray}
SCTC~(Ours) & R50 & \underline{37.92} & \underline{34.78} & {38.86}\\
\rowcolor{lightgray}
SCTC~(Ours) & R101 & {\bf 39.12} & {\bf 36.09} & \underline{39.87}\\
\hline
\end{tabular}
}
\end{center}
\vspace{-0.2in}
\end{table}

\begin{table}[h]
\begin{center}
    \setlength{\abovecaptionskip}{0.1 cm} 
    \setlength{\belowcaptionskip}{-0.2 cm} 
\caption{Performance comparison on V-COCO dataset. 
}
\label{table:v-coco}
\resizebox{\linewidth}{!}
{
\begin{tabular}{llcccc}
\hline
Method & Backbone & $AP_{role}^{\#1}$ & $AP_{role}^{\#2}$ \\
\hline\hline
\emph{One-stage methods} &&&\\
UnionDet~\cite{kim2020uniondet} & R50 &47.5 &56.2 \\
IPNet~\cite{wang2020IPNet} & HG-104 &  51.0 & - \\
HOITrans~\cite{zou2021end} & R101 & 52.9 & - \\
AS-Net~\cite{chen2021reformulating} & R50 &  53.9 & - \\
HOTR~\cite{kim2021hotr} & R50 &  55.2 & 64.4 \\
PhraseHOI~\cite{li2022improving} & R50 & 57.4& -\\
QPIC~(Tamura et al. 2021) & R101 & 58.3 & 60.7 \\
CATN~\cite{dong2022category} & R50  & 60.1 &-\\
GEN-VLKT~\cite{liao2022gen-vlkt} & R50 & 62.4 & 64.5 \\
ParMap~\cite{wu2022mining} &R50 &63.0 &65.1 \\
HOICLIP~\cite{ning2023hoiclip} & R50 & 63.5 & 64.8\\
CDN~\cite{zhang2021mining} & R101 & 63.9 & 65.9 \\
ERNet~\cite{lim2023ernet} & EfficientNetV2-L & 64.2 & -\\
DT~{\cite{zhou2022disentangled}} & R50 & 66.2 & 68.5 \\
\hline
\emph{Two-stage methods} &&&\\
InteractNet~\cite{gkioxari2018detecting} & R50-FPN & 40.0 & 48.0 \\
GPNN~\cite{zhou2021cascaded} & R50 & 50.4 & - \\
ACP~\cite{kim2020cooccur} &R152 & 53.0 & -\\
UPT~(Zhang et al. 2022a) & R101  & 61.3 & 67.1 \\
OpenCat~(Zheng et al. 2023) & R101 & 61.9 & 63.2\\
ViPLO~(Park et al. 2023) & ViT-B/16 & 62.2 & 68.0\\
RmLR~\cite{cao2023re} & R101 & 64.2 & 70.2\\
OCN~\cite{yuan2022detecting} & R101 & 65.3 & 67.1\\
STIP~\cite{zhang2022exploring} & R50 & 66.0 & 70.7 \\
CQL~\cite{xie2023category} & R101 & {66.8} &69.8\\
\rowcolor{lightgray}
SCTC~(Ours) & R50 & \underline{ 67.1} & \underline {71.7} \\
\rowcolor{lightgray}
SCTC~(Ours) & R101 & {\bf 68.2} & {\bf 72.5} \\
\hline
\end{tabular}
}
\end{center}
\vspace{-0.2in}
\end{table}

We report the quantitative results in terms of $AP$ on HICO-DET dataset and V-COCO dataset. 

Tab.~\ref{table:HICO-DET} shows the performance comparison of SCTC and other state-of-the-art methods on HICO-DET dataset.
It can be observed that SCTC outperforms all the existing methods by a large margin.
Specifically, SCTC achieves {\bf 39.12} mAP in the default full setting, obtaining a performance gain of 1.71 mAP (relatively 4.57$\%$) compared to the most recent approach RmLR.
Also, compared with ERNet, which is the 
state-of-the-art one-stage method, our model achieves a significant performance gain of 3.20 mAP (relatively 8.90 $\%$). The results validate the 
superiority of our proposed SCTC.
ACP and STIP directly concatenate the human and object features together and use MLP to fuse them for action prediction, which can't well explore the complex relations between human and object, resulting in relatively worse results.
ViPLO and ParMap explore the potential interactiveness by leveraging body-part features, which guides the detection model to focus on crucial regions of human and object, and have achieved relatively better results.
However, they only consider the relations in the self-triplet level, and ignore the dependencies among different triplets.
For our SCTC, it utilizes the STA to aggregate features from human, object and interaction within each triplet, and then build correlations across different triplets by CTD, which can further promote HOI detection.

For V-COCO, as shown in Tab.~\ref{table:v-coco}, SCTC also performs the best among all the state-of-the-art methods, only using R50 as backbone. 
E.g., SCTC works better than the three recent HOI detectors ERNet, RmLR and CQL  ($AP_{role}^{\#1}$ of 67.1 vs 64.2, 64.2 and 66.8). It is noted that $AP_{role}^{\#2}$ is significantly improved by the SCTC (about 1.9 higher than CQL), and the reason may be that our model can make fully use of scene semantics to infer missed or occluded information.
E.g., when certain parts of humans or objects are occluded, SCTC can rely on the dependencies from different triplets built by the CTD to infer missing information or predict the likely actions based on the available context (please see the supplementary for qualitative results). 

\subsection{Ablation Study}
In this subsection, we explore how Self-Triplet Aggregation (STA), Cross-Triplet Dependency (CTD), and the CLIP-based Knowledge Distillation (KD) affect the HOI detection performance using R50 as backbone.
The ablation studies are conducted on HICO-DET dataset with the ResNet50 backbone, and all training settings are consistent with those mentioned above.

\begin{table}[h]
\begin{center}
\setlength{\tabcolsep}{4pt}
\setlength{\abovecaptionskip}{0.1 cm} 
\setlength{\belowcaptionskip}{-0.1 cm} 
\caption{Effect of STA, CTD, and KD in our model.}
\label{table:main ablation}
\begin{tabular}{ccccccc}
\hline
KD & STA & CTD & MLP & Full & Rare & Non-Rare\\
\hline
\hline
- & - & - & \checkmark & 33.54 & 30.62 & 34.14\\
\checkmark & - & - &\checkmark & 35.63 & 32.82 & 36.35\\
\checkmark & \checkmark & - & - & 36.41 & 32.95 & 37.45\\
\checkmark & - & \checkmark & - & 37.25 & 34.29 & 38.17\\
\checkmark & \checkmark & \checkmark &- & {\bf 37.92} & {\bf 34.78} & {\bf 38.86}\\
\hline
\end{tabular}
\end{center}
\vspace{-0.2in}
\end{table}

\begin{table}[h]
\begin{center}
\setlength{\abovecaptionskip}{0.1 cm} 
\setlength{\belowcaptionskip}{-0.1 cm} 
\caption{Effect of embedding content in STA. IF: interaction feature, SF: spatial feature, LE: learnable edge.}
\label{table:intra content}
\begin{tabular}{cccccc}
\hline
IF & SF & LE & Full & Rare & Non-Rare\\
\hline
\hline
- & - & \checkmark & 35.42 & 32.24 & 36.31\\
- & \checkmark & - & 35.91 & 32.45 & 36.77\\
\checkmark & - & - & 37.76 & 34.83 & 38.71\\
\checkmark & \checkmark & - & {\bf 37.92} & {\bf 34.78} & {\bf 38.86}\\
\hline
\end{tabular}
\end{center}
\vspace{-0.2in}
\end{table}

\subsubsection{Self-Triplet Aggregation.}
SCTC adopts STA to aggregate features from human, object, and interaction. As shown in Tab.~\ref{table:main ablation}, STA with KD (the third row) can improve MLP with KD (the second row) by 0.78 mAP in Full category, validating that our proposed STA is an effective choice to aggregate triplet features. 

The nodes in the graph of STA are initialized with human and object features. Innovatively, we propose to leverage the interaction feature as the edge to connect the nodes.
As shown in Tab.~\ref{table:intra content}, we explore the effect of applying different features to serve as edges to connect the graph nodes.
`Learnable Edge' (LE) means that we directly initialize the edges with learnable parameters without any priors.
It can be observed that the interaction feature (IF) exceeds the LE by 2.34 mAP in the Full category, which proves the effectiveness of IF to connect human and object nodes.
Besides, the spatial feature (SF) also contributes to HOI detection. E.g., IF+SF achieves the best results, improving IF from 37.76 to 37.92 in the Full category.

\begin{table}[h]
\begin{center}
\setlength{\abovecaptionskip}{0.1 cm} 
\setlength{\belowcaptionskip}{-0.1 cm} 
\caption{Effect of Adjacency Matrix in CTD. 
IR: instance relation, SR: semantic relation, LR: layout relation, LE: learnable edge.}
\label{table:CTD}
\begin{tabular}{ccccccc}
\hline
IR. & SR & LR & LE & Full & Rare & Non-Rare\\
\hline
\hline
- & - & - & \checkmark & 36.55 & 33.49 & 37.40\\
\checkmark & - & - & - & 37.14 & 34.38 & 38.04\\
- & \checkmark & - & -& 37.49 & 34.62 & 38.21\\
- & - & \checkmark & - & 37.26 & 34.12 & 38.22\\
\checkmark & \checkmark & \checkmark & - & {\bf 37.92} & {\bf 34.78} & {\bf 38.86}\\
\hline
\end{tabular}
\end{center}
\vspace{-0.2in}
\end{table}

\subsubsection{Cross-Triplet Dependency.}
SCTC adopts CTD to build correlations among different HOI triplets. 
From Tab.~\ref{table:main ablation}, KD+STA+CTD can further improve KD+STA by 1.51 mAP in Full category, which proves the importance to explore relationships across different triplets. 
The nodes in CTD are derived from the STA directly, representing the triplet-level feature.
For the graph edges (i.e.,the adjacency matrix), we initialize them by making full use of instance relation (IR), semantic relation (SR), and layout relation (LR). In Tab.~\ref{table:CTD}, we  
analyze the effect of different relations in the CTD. 
``Learnable edge'' (LE) also means that the edges are initialized with learnable parameters without any priors.
It can be found that any one of the IR, SR, and LR works better than the LE by 0.59 mAP, 0.94 mAP, and 0.71 mAP in the Full category, respectively. Besides, when jointly considering IR, SR and LR, the HOI detection can be further improved.

\begin{table}[h]
\begin{center}
\setlength{\abovecaptionskip}{0.1 cm} 
\setlength{\belowcaptionskip}{-0.1 cm} 
\caption{Effect of 
Knowledge Distillation (KD) strategies.}
\label{table:kd teacher}
\begin{tabular}{cccccc}
\hline
CLIP & Bert & CE  & Full & Rare & Non-Rare\\
\hline
\hline
\ - & - & - & 35.95 & 32.57 & 37.01\\
\checkmark & - & - & {\bf 37.92} & {\bf 34.78} & {\bf 38.86}\\
- & \checkmark & - & 35.78 & 32.25 & 36.69\\
- & - & \checkmark & 37.36 & 34.21 & 38.41\\
\checkmark & - & \checkmark & 37.79 & 34.61 & 38.66\\
\hline
\end{tabular}
\end{center}
\vspace{-0.2in}
\end{table}

\subsubsection{Knowledge distillation.}
We transfer the knowledge from CLIP to our SCTC for the refinement of interaction feature. Specifically, we use the text encoder to generate text embeddings to distill our interaction feature.
From Tab.~\ref{table:main ablation}, KD+MLP improves MLP by 2.09 mAP, which validates the significance of knowledge distillation to enhance interaction feature.
Here, we also explore other three 
strategies to distill the interaction feature as shown in Tab.~\ref{table:kd teacher}.
`Bert' means that we leverage Bert model ~\cite{devlin-etal-2019-bert} to generate text embeddings, and `CE' means that we directly use interaction feature to generate action classification predictions, supervised by ground truth action labels. 
It can be observed that CLIP achieves the better results than those of the Bert and CE. 
Bert model lacks the alignment of visual feature and linguistic feature, which results in the incompatibility between the interaction feature and its text embeddings.
Besides, the interaction feature contains integrated information of human, object, and interaction, so it is inappropriate to distill it only with the action labels.
This is also the reason why CLIP+CE performs worse than CLIP. Differently, CLIP model can fully encode the {\textless human, object, action\textgreater} triplet and its textual embedding is compatible with the visual feature, so it is suitable to distill the interaction feature.

\section{Conclusion}
In this paper, we propose the SCTC, the state-of-the-art two-stage HOI detector, to jointly explore Self- and Cross-Triplet Correlations for HOI detection.
For the Self-Triplet Aggregation, 
SCTC creates a graph for each candidate triplet to jointly aggregate instance-level features from human, object and the potential action semantics.
As for the Cross-triplet Dependency, it builds the contextual dependencies across different triplet proposals, where the adjacency matrix is formulated as the combination of instance relation, semantic relation and layout relation. Besides, we transfer the text knowledge from CLIP to SCTC, which helps to enhance the representation of interaction features. 
Experimental results on HICO-DET and VCOCO datasets verify the effectiveness of the proposed SCTC.

\bigskip
\section{Acknowledgments}
This work was supported in part by the National Key Research and Development Program of China under Grant 2022YFB4703201, in part by the National Natural Science Foundation of China under Grants 62206075, 61733011, and 62261160652, in part by the GuangDong Basic and Applied Basic Research Foundation under Grant 2021A1515110438, and in part by the Shenzhen Science and Technology Program under Grant RCBS20221008093220004.

\bibliography{ref}
\clearpage
\end{document}